\documentclass[lettersize,journal]{IEEEtran}

\usepackage{amsmath,amsfonts,amssymb}  
\usepackage{bm}                         
\usepackage{bbm}                        
\usepackage{graphicx}                   
\usepackage{stfloats}                   
\usepackage{subfig}                     
\usepackage{multirow}                   
\usepackage{booktabs}                   
\usepackage{hyperref}                   
\usepackage{booktabs}
\usepackage{hyperref}
\usepackage{cleveref}
\usepackage{tikz}
\usepackage{adjustbox}
\usepackage{amsthm}
\usepackage{subfig}

\newtheorem{example}{Example}

\title{VT-Former: Efficient Transformer-based Decoder for Varshamov-Tenengolts Codes}

\author{Yali~Wei, 
        Alan~J.X.~Guo,~\IEEEmembership{Member,~IEEE,} 
        Zihui~Yan, 
        Yufan~Dai,
        Wenjia~Fan
\thanks{This work was supported by the National Key Research and Development Program of China under Grant 2025YFC3409900, 
the National Natural Science Foundation of China, 
and the Emerging Frontiers Cultivation Program of Tianjin University Interdisciplinary Center. (Corresponding author: Alan J.X. Guo.)}
\thanks{Y. Wei, Y. Dai, and W. Fan are with 
Center for Applied Mathematics, KL-AAGDM, Tianjin University, Tianjin 300072, China. 
A. Guo is with 
Center for Applied Mathematics, KL-AAGDM, and State Key Laboratory of Synthetic Biology, 
Tianjin University, Tianjin 300072, China. 
Z. Yan is with the Frontiers Science Center for Synthetic Biology 
and Key Laboratory of Systems Bioengineering (Ministry of Education), 
School of Chemical Engineering and Technology, and State Key Laboratory of Synthetic Biology, 
Tianjin University, Tianjin 300072, China. }

}
\begin{document}

\maketitle

\begin{abstract}
In recent years, widespread attention has been drawn to the challenge of correcting 
insertion, deletion, and substitution (IDS) errors in DNA-based data storage. 
Among various IDS-correcting codes, Varshamov-Tenengolts (VT) codes, 
originally designed for single-error correction, have been established as a central research focus. 
While existing decoding methods demonstrate high accuracy for single-error correction, 
they are typically not applicable to the correction of multiple IDS errors. 
In this work, the latent capability of VT codes for multiple-error correction is investigated through 
a statistic-enhanced Transformer-based VT decoder (VT-Former), 
utilizing both symbol and statistic feature embeddings.
Experimental results demonstrate that VT-Former achieves nearly 100\% accuracy on correcting single errors. 
For multi-error decoding tasks across various codeword lengths, 
improvements in both frame accuracy and bit accuracy are observed, 
compared to conventional hard-decision and soft-in soft-out decoding algorithms. 
Furthermore, while lower decoding latency is exhibited by the base model compared to traditional soft decoders, 
the architecture is further optimized in this study to enhance decoding efficiency and reduce computational overhead.
\end{abstract}


\section{Introduction}

In recent years, DNA has garnered substantial attention from both academia and industry as 
a promising information storage medium due to its longevity and 
high storage density~\cite{goldman2013towards,erlich2017dna,sini2019reconstruction,song2022robust}. 
One of the central challenges in DNA data storage systems is the correction of sequence errors introduced 
during synthesis and sequencing processes, particularly insertion, deletion, and 
substitution (IDS) errors~\cite{yazdi2015dna,heckel2019characterization,li2023multiple}. 
Such errors not only directly compromise data integrity but can also pose serious threats to the overall 
reliability of the storage system and the stability of data retrieval. Therefore, developing efficient and 
robust methods for correcting IDS errors has become a key research topic in the field of DNA-based data storage. 

To correct IDS errors, several coding schemes have been proposed, 
including convolutional codes~\cite{maarouf2022concatenated,buttigieg2015improved,press2020hedges}, 
watermark codes~\cite{davey2001reliable}, time-varying block codes~\cite{buttigieg2014time}, 
and Varshamov-Tenengolts (VT) codes~\cite{levenshtein1965binary}. 
Among these, VT codes are regarded as an effective error correction scheme for DNA storage owing 
to their asymptotically optimal redundancy efficiency. 
Specifically, the VT code requires only $\lceil \log n \rceil + 1$ 
redundant bits to correct a single IDS error, offering high theoretical coding efficiency. When a single IDS error occurs 
in a codeword, conventional hard-decision (HD) decoding methods can effectively recover the 
original information~\cite{levenshtein1965binary,varsamov1965code,cai2021correcting}. 
However, in practical applications, multiple random IDS errors are often encountered, 
leading to a significant degradation in the performance of HD decoding and consequently impairing 
both decoding accuracy and system utility.

To improve decoding performance in the presence of multiple errors, 
a soft-in soft-out (SISO) decoding algorithm~\cite{yan2023novel} was proposed, 
which enhances decoding robustness by incorporating symbol-level posterior probability information. 
Compared with HD decoding, the SISO method exhibits better error correction capability when handling multiple errors. 
Nevertheless, although SISO achieves some improvement in bit-level accuracy, its enhancement in frame-level 
decoding success remains limited. Moreover, as the number of errors increases or the codeword length grows, 
the computational complexity of the SISO method rises sharply, making it difficult to meet the real-time decoding 
speed requirements of practical systems.

Motivated by the success of the Transformer~\cite{vaswani2017attention} in error correction for 
linear codes~\cite{choukroun2022error,park2025crossmpt}, 
a statistic-enhanced Transformer-based VT decoder (VT-Former) is proposed in this paper 
to advance soft decoding capabilities for VT code. 
Experimental results demonstrate that, compared with traditional methods, 
VT-Former not only improves multi-error correction performance but also accelerates 
the decoding process through GPU-based parallel computing. 

The main contributions of this paper are summarized as follows:
\begin{itemize}
    \item \textbf{First deep learning-based decoder for VT codes:} 
    A codeword embedding method integrating symbolic and statistic features is developed, 
    facilitating the application of a Transformer-based model as a soft decoder. 
    Furthermore, the efficacy of various Transformer architectures and strategies 
    for IDS error correction in VT codes is evaluated. 
    \item \textbf{Performance improvement in multi-error correction scenarios:} 
    Across various codeword lengths, VT-Former achieves lower frame error rates compared to existing methods, 
    with reductions ranging from 0.1\% to 15.4\%. 
    Furthermore, the bit error rate is decreased by 0.6\% to 4.3\% through the incorporation of a window masking mechanism.
    \item \textbf{Accelerated decoding speed:} 
    Experimental results demonstrate that a speedup of over $5\times$ is achieved by the VT-Former series 
    compared to SISO decoding. 
    Moreover, for long codewords, a speedup exceeding $44\times$ is observed, 
    whereby the practicality of the model in large-scale DNA data storage systems is substantially enhanced. 

\end{itemize}

\section{Related Work}

The decoding of VT codewords with IDS errors has long been a central research topic in the field of information coding, 
particularly in terms of decoding accuracy and efficiency. 
To address this issue, 
a combinatorial HD method based on the weights and checksums of the 
received sequence was proposed~\cite{levenshtein1965binary}. 
This method accurately locates errors by evaluating the weights of the received sequence and 
combining the checksums for correction, enabling the effective detection and correction of single IDS errors. 
However, when multiple errors occur in VT codewords, this HD decoding method experiences a significant decline in accuracy.

To facilitate information recovery in the presence of multiple errors, 
a SISO decoding strategy was proposed in~\cite{yan2023novel}.
This method relies on the Bahl-Cocke-Jelinek-Raviv (BCJR) algorithm and 
adopts a bitwise maximum a posteriori (MAP) decoding strategy. 
Unlike traditional HD decoding, the capability to address multiple-error correction was first introduced by this SISO approach. 
Nevertheless, as the number of errors increases, the decoding accuracy is observed to decline, 
and the high computational complexity limits its practical applicability. 

Error correcting codes (ECC) improve data reliability by adding redundant bits, 
enabling the detection and correction of errors during transmission or storage. Common ECCs include Hamming codes, 
BCH codes, LDPC codes, and Turbo codes, 
which are widely used to ensure data integrity~\cite{kim2015error}. 
In recent years, the rapid development of deep learning has brought revolutionary improvements to ECC, 
where belief propagation (BP) decoders based on neural networks have gradually 
become a mainstream solution~\cite{nachmani2019hypergraph}. 
Compared to traditional decoding methods, neural BP decoders not only improve decoding efficiency 
but also enhance robustness to noise~\cite{lugosch2017neural,nachmani2016learning}. 

Considering the outstanding performance of Transformer models in various tasks~\cite{vaswani2017attention}, 
they have also been introduced into ECC research. 
In~\cite{choukroun2022error,choukroun2024foundation,choukroun2024learning,park2025crossmpt},  
this field is advanced by utilizing Transformer models for decoding linear codewords. 
Their research demonstrated that Transformers improve decoding accuracy while reducing computational complexity, 
significantly surpassing existing state-of-the-art decoding methods. 

However, despite the success of Transformer models in linear codes, 
their application in VT code decoding remains a relatively unexplored research area.

\section{Background}

This study focuses on decoding binary VT codewords. 
The following provides background on VT codes and related decoding algorithms for correcting insertion, 
deletion, and substitution (IDS) errors. 

\subsection{VT Codes}
VT codes are a class of algebraic block codes consisting of all binary vectors of length $n$ that satisfy the following condition:
\begin{equation}\label{eqn:vtformula}
\mathrm{VT}_{a, m}(n) = \Big\{ v \in \{0,1\}^n : \sum_{i=1}^n i \, v_i \equiv a \pmod{m} \Big\},
\end{equation}
where $m$ is a predefined integer, and $a \in \{0, 1, \dots, m-1\}$ is the parity of the sequence $v$. When $m \ge n+1$, the code can correct a single insertion or deletion error, and when $m \ge 2n$, it can correct a single IDS error \cite{levenshtein1965binary}. In this work, we focus on VT codes that correct IDS errors with $m = 2n + 1$.

\subsection{VT Encoder}\label{sec:vtencoder}
A systematic encoding scheme for VT codes \cite{saowapa1999systematic}, referred to as the VT encoder, is summarized as follows.  
Given a message sequence $u = (u_1, u_2, \dots, u_p) \in \{0,1\}^p$, the encoder generates a codeword $v = \mathrm{VT}(u) \in \mathrm{VT}_{a, 2n+1}(n)$, where $y = n - \lceil \log n \rceil - 1$. Parity check bits are inserted at positions $v_{2^i}$ for $0 \le i \le n-y-2$, and at $v_n$, ensuring
\begin{equation}\label{eqn:vtencoderformula}
\sum_{i=1}^{n} i \, v_i \equiv a \pmod{2n+1}.
\end{equation}

\begin{example}
Given a message sequence $u = 11011$ and parity $a = 0$, we have $n = 10$, $y = 5$, and $m = 21$. The codeword $v = (v_1, v_2, \dots, v_{10})$ must satisfy
\begin{equation}
\sum_{i=1}^{10} i \, v_i \equiv \sum_{i=1}^{4} 2^{i-1} v_{2^{i-1}} + 10 \, v_{10} \equiv 0 \pmod{21}.
\end{equation}
Solving, we obtain $v_{10} = 1$, $v_8 = 1$, and $v_1 = v_2 = v_4 = 0$. Expanding $18-10 = 8$ in binary form $2^3$, the final codeword is $\bar{0}\,\bar{0}\,1\,\bar{0}\,1\,0\,1\,\bar{1}\,1\,\bar{1}$, where overlined positions denote parity bits.
\end{example}

\subsection{Hard-Decision Decoder}
The HD decoder detects and corrects a single error in a codeword. The procedure is as follows:
\begin{itemize}
    \item Compute the parity check for the received sequence $r = (r_1, r_2, \dots, r_N)$:
    \begin{equation}
        \overline{a} = \sum_{i=1}^{N} i \, r_i \pmod{2n + 1}.
    \end{equation}
    If $\overline{a} \ne a$, the codeword is corrupted.

    \item Represent the corrupted sequence as $r = [v_1, \dots, v_{t-1}, p, v_t, \dots, v_n]$, 
    where $p$ is the error symbol. Compute the difference 
    \begin{equation}
        \overline{a} - a \equiv \sum_{i=j}^{n} v_i + j \, p \pmod{2n+1}.
    \end{equation}
    This guides the correction.

    \item If $\overline{a} - a \le \sum_{i=1}^{N} r_i$, then $p = 0$, and the insertion occurs before the $(\overline{a}-a)$-th "1" (counted from back to front). Otherwise, $p = 1$, and its position is after the $(\overline{a}-a-\sum r_i + 1)$-th "0". Substitution and deletion errors are treated similarly.
\end{itemize}

\subsection{SISO Decoder}
The SISO decoder uses bitwise MAP decoding based on the BCJR algorithm. For a received sequence $r = (r_1, r_2, \dots, r_N)$, the posterior probability of a bit sequence $v = (v_1, \dots, v_n)$ is computed using aggregated states $S_t$ and drift values $D_t$:
\begin{align}
&\Pr[v_{1}^{n} \mid r_{1}^{N}, S_0=s_0, D_0=d_0] \, \Pr[r_{1}^{N} \mid S_0=s_0, D_0=d_0] \nonumber\\
&= \prod_{t=1}^{n} \mathcal{F}[v_t, r_{d_{t-1}+t}^{d_t+t}, s_t, d_t],
\end{align}
where
\begin{equation}
\begin{split}
\mathcal{F}[v_t, r_{d_{t-1}+t}^{d_t+t}, s_t, d_t] 
&=  \Pr\big[r_{d_{t-1}+t}^{d_t+t} \mid v_t, s_t, s_{t-1}, d_t, d_{t-1}\big] \\
& \times \Pr[v_t] \, \times \Pr[s_t, d_t \mid s_{t-1}, d_{t-1}].
\end{split}
\end{equation}

The log-posterior probability (log-APP) for each bit $v_t$ is
\begin{align}
L(v_t) &= \log \frac{\Pr[v_t = 0 \mid r, \mathcal{S}]}{\Pr[v_t = 1 \mid r, \mathcal{S}]} \nonumber\\
&= \log \frac{\Pr[r, \mathcal{S} \mid v_t = 0]}{\Pr[r, \mathcal{S} \mid v_t = 1]} + \log \frac{\Pr[v_t = 0]}{\Pr[v_t = 1]},
\end{align}
where $\mathcal{S}$ represents the aggregated state. The recursion relations for computing the forward and backward probabilities are
\begin{align}
\alpha_t(s,d) &= \sum_{(s',s), (d',d)} \gamma_t(s',d',s,d)\,\alpha_{t-1}(s',d'),\\
\beta_t(s',d') &= \sum_{(s',s), (d',d)} \gamma_{t+1}(s',d',s,d)\,\beta_{t+1}(s,d),
\end{align}
with the state transition probability
\begin{align}
\gamma_t(s',d',s,d) &= \Pr[r_{d'+t}^{d+t}, s, d \mid s', d'] \nonumber\\
&= \Pr[s \mid s'] \cdot \Pr[r_{d'+t}^{d+t}, d \mid s', d', s].
\end{align}

Finally, a hard decision is made based on the sign of $L(v_t)$:
\begin{equation}
\hat{v_t} = \operatorname{sign}(L(v_t)).
\end{equation}
The original message is then recovered via inverse encoding.

\section{Method}
In this section, the overall framework of the proposed decoder is presented, 
including the symbol- and statistic-based codeword embedding and the attention masking strategy. 

\subsection{Framework}
As illustrated in~\Cref{fig:model}, the proposed VT-Former model is constructed based on a 
Transformer-based encoder-decoder architecture\footnote{In this context, 
the terms `encoder' and `decoder' are used to denote the specific components of the Transformer architecture, 
rather than the error-correction codecs.}. 
The model adopts a streamlined encoding design, in which the encoder is directly formed by the embedding 
representation of the corrupted codeword. This embedding integrates both symbolic information and statistic features. 
The decoder employs an autoregressive symbol-by-symbol prediction mechanism, using the codeword embedding as external 
memory and gradually reconstructing the original codeword through a multi-layer attention network.

{\color{black}
To further investigate the impact of local versus global contextual information on decoding performance, 
a combined masking strategy is introduced within the decoder. 
Specifically, an adjustable local window mask is applied to the decoder's self-attention mechanism, 
modifying the standard causal upper-triangular mask used in the sequence-to-sequence module. 
When the window size is set to $w=1$, the model attends exclusively to the immediately preceding symbol, 
thereby prioritizing local context. 
Conversely, setting the window size equal to the code length ($w=n$) allows the model to attend to 
all previously generated symbols, effectively capturing the global context.


During the training phase, the teacher forcing strategy is adopted, 
where the ground-truth codeword, processed with masking and positional shifting, serves as the decoder input. 
The model parameters are optimized using a cross-entropy loss function. 
In the testing phase, an autoregressive generation approach is employed, where the model uses the predictions 
from previous steps as the current input to progressively reconstruct the entire codeword.}


\begin{figure}[t!]
    \centering
    \usetikzlibrary{shapes.geometric, arrows.meta, positioning, calc, fit}
    \usetikzlibrary{backgrounds}
    \definecolor{colorEmbed}{RGB}{235, 235, 250}   
    \definecolor{colorMHA}{RGB}{255, 235, 210}     
    \definecolor{colorFFN}{RGB}{230, 240, 255}     
    \definecolor{colorAddNorm}{RGB}{250, 250, 210} 
    \definecolor{colorHead}{RGB}{220, 240, 220}    
    \definecolor{colorgray}{RGB}{250, 250, 250}    
\begin{adjustbox}{width=0.9\linewidth}
    \begin{tikzpicture}[
        font=\small,
        >=Stealth,
        block/.style={draw, rectangle, minimum width=2.5cm, minimum height=0.6cm, rounded corners=3pt, align=center, line width=0.6pt},
        arrow/.style={->, line width=0.7pt},
        residual/.style={->, line width=0.6pt, rounded corners=5pt}
    ]

        \node (input1) at (0,0) {Corrupted Codewords};
        
        \node[block, fill=colorEmbed, above=0.5cm of input1] (stat) {Statistic Embedding};
        \node[block, fill=colorEmbed, above=0.45cm of stat] (sym) {Symbol Embedding};
        \node at ($(stat)!0.5!(sym)$) {\large $\oplus$};
        \node[draw, dashed, inner sep=5pt, rounded corners=5pt, fit=(stat) (sym)] (embed_box1) {};
        
        \draw[arrow] (input1) -- (embed_box1);
        
        \node[block, fill=colorMHA, above=0.7cm of embed_box1] (mha1) {MHA};
        \node[block, fill=colorAddNorm, above=0.3cm of mha1] (an1) {Add \& Norm};
        \node[block, fill=colorFFN, above=0.5cm of an1] (ff1) {Feed Forward};
        \node[block, fill=colorAddNorm, above=0.3cm of ff1] (an2) {Add \& Norm};
        
        \draw[arrow] (embed_box1) -- (mha1);
        \draw[arrow] (mha1) -- (an1);
        \draw[arrow] (an1) -- (ff1);
        \draw[arrow] (ff1) -- (an2);

        \node[left=0.5cm of an1] (nx) {\large $N_{\mathrm{en}} \times$};
        
        \draw[residual] (embed_box1.north) -- ++(0,0.3) -- ++(-1.6,0) |- (an1.west);
        \draw[residual] (an1.north) -- ++(0,0.2) -- ++(-1.6,0) |- (an2.west);
        \draw[residual] ($(mha1.south) + (0,-0.3)$) -- ++ (-1,0) -- ($(mha1.south) + (-1,0)$); 
        \draw[residual] ($(mha1.south) + (0,-0.3)$) -- ++ (1,0) -- ($(mha1.south) + (1,0)$); 

        \begin{scope}[on background layer]
            \coordinate (enc_box_west) at ($(mha1.west)+(-0.35,0)$);
            \coordinate (enc_box_south) at ($(mha1.south)+(0,-0.4)$);
            \node[draw, inner sep=5pt, rounded corners=5pt, fit=(mha1) (an2) (enc_box_west) (enc_box_south), fill=gray, fill opacity=0.1] (enc_box) {};
        \end{scope}
        \begin{scope}[shift={(3.7,0)}]
            \node (input2) at (0,0) {Codewords (Shifted right)};
            \node[block, fill=colorEmbed, above=0.5cm of input2] (embed2) {Embedding};
            
            \draw[arrow] (input2) -- (embed2);
            
            \node[block, fill=colorMHA, above=0.7cm of embed2] (mha2) {MHA};
            \node[block, fill=colorAddNorm, above=0.3cm of mha2] (an3) {Add \& Norm};
            \node[block, fill=colorMHA, above=0.6cm of an3] (mha3) {MHA};
            \node[block, fill=colorAddNorm, above=0.3cm of mha3] (an4) {Add \& Norm};
            \node[block, fill=colorFFN, above=0.4cm of an4] (ff2) {Feed Forward};
            \node[block, fill=colorAddNorm, above=0.3cm of ff2] (an5) {Add \& Norm};
            
            \draw[arrow] (embed2) -- (mha2);
            \draw[residual] ($(mha2.south) + (0,-0.3)$) -- ++ (-1,0) -- ($(mha2.south) + (-1,0)$); 
            \draw[residual] ($(mha2.south) + (0,-0.3)$) -- ++ (1,0) -- ($(mha2.south) + (1,0)$); 
            \draw[arrow] (mha2) -- (an3);
            \draw[residual] (an3) -- ($(mha3.south)+(0,-0.3)$) -- ($(mha3.south)+(1,-0.3)$) -- ($(mha3.south)+(1,0)$);
            \draw[arrow] (mha3) -- (an4);
            \draw[arrow] (an4) -- (ff2);
            \draw[arrow] (ff2) -- (an5);
            
            \draw[residual] (embed2.north) -- ++(0,0.3) -- ++(1.6,0) |- (an3.east);
            \draw[residual] ($(an3.north)+(0,0.1)$) -- ++(1.6,0) |- (an4.east);
            \draw[residual] ($(an4.north)+(0,0.1)$) -- ++(1.6,0) |- (an5.east);

            
            \node[right=0.5cm of an4] (nx) {\large $N_{\mathrm{de}} \times$};
            \begin{scope}[on background layer]
                \coordinate (dec_box_east) at ($(mha2.east)+(0.35,0)$);
                \coordinate (dec_box_south) at ($(mha2.south)+(0,-0.4)$);
                \node[draw, inner sep=5pt, rounded corners=5pt, fit=(mha2) (an5) (dec_box_east) (dec_box_south),fill=gray,fill opacity=0.1] (dec_box) {};
            \end{scope}
            \node[block, fill=colorHead, above=0.5cm of an5] (head) {Prediction Head};
            \draw[arrow] (an5) -- (head);
        \end{scope}

        \draw[residual] (an2.north) -- ++(0,0.5) -- ++(1.85,0) |- ($(mha3.south)+(0,-0.3)$) -- (mha3.south);
        \draw[residual] (an2.north) -- ++(0,0.5) -- ++(1.85,0) |- ($(mha3.south)+(-1,-0.3)$) -- ($(mha3.south)+(-1,0)$);
    \end{tikzpicture}
\end{adjustbox}
    \caption{Framework of the VT-Former architecture. 
    A sequence-to-sequence model is employed with symbol- and statistic-based embedding. 
    The decoder adopts a next-symbol prediction scheme enhanced by a combined masking strategy.}
    \label{fig:model}
\end{figure}

\subsection{Symbol- and Statistic-based Codeword Embedding}\label{sec:symbol-statistic}
It is observed that using only the symbol embedding of a codeword results in poor correction accuracy, 
as presented in~\Cref{tab:statistic}. 
Motivated by the HD decoder, the statistics of the symbol positions are integrated into the codeword embedding in 
the proposed method. Both received/corrupted codewords and transmitted/ground-truth codewords are embedded in the same manner. 
For each VT codeword, the following operations are performed to help the model capture its feature information. 

To be precise, let $\bm{c} = (c_1, c_2, \dots, c_n), \; c_i\in\{0,1\}$ be a codeword of length $n$. The codeword embedding is defined as
\begin{equation}
    \Phi(\bm{c}) = \mathrm{concat}\big(\Phi_{\mathrm{stat}}(\bm{c}),\; \Phi_{\mathrm{sym}}(\bm{c})\big),
\end{equation}
which is a concatenation of the symbol embedding $\Phi_{\mathrm{sym}}(\bm{c})$ and the statistic embedding $\Phi_{\mathrm{stat}}(\bm{c})$.

\subsubsection{Symbol Embedding of a Codeword}

\begin{figure}[t!]
    \centering
\begin{adjustbox}{width=0.95\linewidth}
\begin{tikzpicture}[node distance=0cm, font=\small]
    \definecolor{highlightColor}{RGB}{230, 240, 255} 

    \foreach \i in {0,...,4} {
        \foreach \j in {0,1} {
            \node[draw, minimum width=0.7cm, minimum height=1cm, rounded corners=2pt] (m\i\j) at (\i*0.7, -\j*1) {$\bm{e}_{\i\j}$};
        }
    }
    \fill[highlightColor, opacity=1] (m00.north west) rectangle (m00.south east);
    \fill[highlightColor, opacity=1] (m11.north west) rectangle (m11.south east);
    \fill[highlightColor, opacity=1] (m20.north west) rectangle (m20.south east);
    \fill[highlightColor, opacity=1] (m31.north west) rectangle (m31.south east);
    \fill[highlightColor, opacity=1] (m41.north west) rectangle (m41.south east);

    \foreach \i/\j in {0/0, 1/1, 2/0, 3/1, 4/1} {
        \node[draw, thick, minimum width=0.7cm, minimum height=1cm, rounded corners=2pt] at (\i*0.7, -\j*1) {$\bm{e}_{\i\j}$};
    }

    \node at (1.37, -1.87) {$\bm{v} = (0, 1, 0, 1, 1)$};

    \draw[-stealth, thick] (3.3, -0.5) -- (4.5, -0.5) node[midway, above] {\small Lookup};

    \begin{scope}[shift={(5.0, -0.5)}]
        \foreach \i/\idx in {0/00, 1/11, 2/20, 3/31, 4/41} {
            \node[thick, draw, fill=highlightColor, minimum width=0.7cm, minimum height=1.cm, rounded corners=2pt] at (\i*0.7, 0) {$\bm{e}_{\idx}$};
        }
        \node at (1.4, 0.8) {$\Phi_{\mathrm{sym}}(\bm{v})$};
        \node at (1.4, -0.9) {Symbol Embedding};
    \end{scope}
\end{tikzpicture}
\end{adjustbox}
    \caption{Symbol embedding of a codeword. For a codeword $\bm{v} = (0,1,0,1,1)$, the vectors $(\bm{e}_{00},\bm{e}_{11},\bm{e}_{20},\bm{e}_{31},\bm{e}_{41})$ are gathered to form the symbol embedding of the codeword.}
    \label{fig:symbol-embedding}
\end{figure}

In general, the embedding of sequences in a sequence-to-sequence model consists of word embeddings 
and positional embeddings, which are typically added or concatenated. 
The positional embedding encodes the positional information of the words and can be either fixed or learnable.

In this work, due to the simplicity of the binary symbols in the codeword, 
the learnable positional embedding is integrated into the symbol embedding. 
Specifically, as illustrated in~\Cref{fig:symbol-embedding}, 
two learnable embedding vectors are associated with each position of the codeword, 
representing the binary symbols $0$ and $1$, respectively. 
The symbol embedding $\Phi_{\mathrm{sym}}(\bm{c}) = (\phi_1, \phi_2 , \dots , \phi_n)$ for codeword $\bm{c}$ is formulated as
\begin{equation}
    \phi_i = (1-c_i)\bm{e}_{i0} + c_i\bm{e}_{i1}, 
\end{equation}
where $\bm{e}_{i0}$ and $\bm{e}_{i1}$ are the learnable embedding vectors associated with position $i$. 


\subsubsection{Statistic Embedding of a Codeword}
The VT code corrects IDS errors using redundancy information based on joint 
symbol-position constraints, as described in~\Cref{eqn:vtformula} and \Cref{sec:vtencoder}. 
Motivated by this, the joint statistics of position and symbol are embedded as part of the final codeword embedding.

Analogous to \Cref{eqn:vtencoderformula}, the two statistics that sum the position indices of symbol `$0$'s 
and the position indices of symbol `$1$'s, respectively, are computed as
\begin{align}
    s_0 &= \sum_{i=1}^{n} i \cdot \mathbbm{1}_{\{c_i = 0\}}, \\
    s_1 &= \sum_{i=1}^{n} i \cdot \mathbbm{1}_{\{c_i = 1\}},
\end{align}
where $\mathbbm{1}_{X}$ denotes the indicator function of set $X$.

Two lookup tables $\phi_{\mathrm{0}}, \phi_{\mathrm{1}}$ are used to store the embedding vectors for 
the possible values of the two statistics. The statistic embedding of the codeword is then obtained by 
retrieving the embedding vectors $\Phi_{\mathrm{stat}}=(\phi_{\mathrm{0}}(s_0), \phi_{\mathrm{1}}(s_1))$ from 
the tables according to $s_0$ and $s_1$, respectively.

\subsection{Combined Masking Strategy}
In sequence-to-sequence models, the standard causal masking mechanism guarantees autoregressive decoding 
but often fails to adequately attend to local context information. 
To enhance bit-level prediction accuracy, this work adopts a combined masking strategy that integrates a local window mask. 
This approach adheres to the autoregressive constraints of the upper-triangular mask while explicitly modeling dependencies 
between neighboring symbols through an adjustable sliding window. 

\subsubsection{Upper Triangular Mask}
Define the upper triangular mask $M_{\mathrm{upper}} \in \mathbb{R}^{L \times L}$, where $L$ represents the 
length of the codeword embedding sequence:
\begin{equation}
M_{\mathrm{upper}}(k, j) = 
\begin{cases} 
0, & k \leq j, \\ 
-\infty, & k > j,
\end{cases}
\end{equation}
where $k$ and $j$ correspond to the row and column indices of the attention weight matrix, respectively.

\subsubsection{Window Mask}
The local window attention mechanism originally stems from the Swin Transformer \cite{liu2021swin}, whose core idea is to confine attention computation to a limited local neighborhood. Drawing on this concept, this paper constructs a window mask \( M_{\mathrm{window}} \in \mathbb{R}^{L \times L} \), which restricts each position to attend only to a local contextual region centered around it with a window radius \( w \). The specific definition is as follows:
\begin{equation}\label{eqn:windowmask}
    M_{\mathrm{window}}(k, j) =
    \begin{cases}
        0, & \text{if } |k - j| \leq w, \\
        -\infty, & \text{otherwise}.
    \end{cases}
\end{equation}
By adjusting the window size \( w \), the scope of local context available during decoding can be flexibly controlled.

\subsubsection{Mask Fusion Mechanism}
During model training, the mask applied in attention computation is the superposition of the two masks, i.e.,
\begin{equation}
    M = M_{\mathrm{upper}} + M_{\mathrm{window}}.
\end{equation}
The attention weight calculation is then
\begin{equation}
    A_H(Q, K, V) = \mathrm{Softmax} \Big( \frac{Q K^T}{\sqrt{d}} + M \Big) V,
\end{equation}
where \(Q, K, V \in \mathbb{R}^{L \times d}\) are the query, key, and value matrices, and \(d\) is the 
embedding dimension \cite{vaswani2017attention}.

During decoding inference, the model iteratively predicts subsequent symbols using the local segment \(\bm{c}[-w:]\) of the last \(w\) symbols of the currently generated codeword, combined with the embedded representation of the corrupted codeword.

\section{Experimental Evaluation}
\subsection{Experimental Setup}
To ensure representativeness and statistical significance, this study evaluates performance on codewords of 
three lengths: 20 bits (short), 68 bits (medium), and 100 bits (long). 
For a given codeword length, the corresponding message length is determined by \Cref{eqn:vtencoderformula}. 
The three VT codes under consideration are denoted by their (codeword, message) lengths as
$\mathrm{VT}(20,14)$, $\mathrm{VT}(68,60)$, and $\mathrm{VT}(100,92)$ respectively. 

To ensure a rigorous separation between the training and testing data, 
a deterministic hash-based dataset partitioning strategy is employed. 
Let $M = \{0,1\}^k$ denote the set of message sequences, the training and testing sets, 
$\mathcal{D}_{\mathrm{train}}$ and $\mathcal{D}_{\mathrm{test}}$, 
are constructed as follows: 
\begin{eqnarray}
    \mathcal{D}_{\mathrm{train}} = \{\mathrm{VT}(\bm{m})|\, \mathrm{hash}(\bm{m}) \bmod 11 \neq 0, \forall \bm{m} \in M\}; \\
    \mathcal{D}_{\mathrm{test}} = \{\mathrm{VT}(\bm{m})|\, \mathrm{hash}(\bm{m}) \bmod 11 = 0, \forall \bm{m} \in M\}.
\end{eqnarray}
Subsequently, random samples drawn from these respective datasets are utilized for model training and evaluation. 

The learning rate was set to $0.0001$ and adjusted using a cosine annealing scheduler \cite{loshchilov2017sgdr}. 
The embedding dimension is 512, and the feedforward network dimension is four times 
the embedding dimension \cite{Ruibin2020OnLayerNormalization}. 
The VT-Former consists of $N_{\mathrm{en}}=3$ encoder layers and $N_{\mathrm{de}}=3$ decoder layers, each with $8$ attention heads. 
The default window mask size corresponds to the codeword length, indicating that no masking is applied. 
Experiments with specific window sizes are denoted by a numerical suffix, for instance, 
VT-Former-1 signifies a window size of $1$, where bits are predicted independently.





To comprehensively evaluate the performance of the proposed method, 
both the bit error rate (BER) and frame error rate (FER) are employed. 
The BER represents the fraction of erroneous bits in the decoded codewords, 
while the FER is the ratio of codewords corrected in error to the total number of corrupted codewords received. 
It is worth noting that since the codewords are binary and randomly generated, the BER cannot exceed $50\%$. 

\subsection{Comparative Experiments}
To systematically evaluate the decoding performance of the proposed method, 
a comparative analysis is conducted between VT-Former and two baseline algorithms: 
the HD decoder and the SISO decoder.  
The experiments are carried out under two distinct task settings: 
the correction of a fixed number of random IDS errors, and 
the correction of randomly introduced IDS errors governed by a specific channel error rate. 
It is noteworthy that the HD decoder declares a decoding failure whenever the number of errors exceeds 
its error-correcting capability. 
Similarly, the SISO decoder is considered to have failed if the iterative process 
reaches a predefined maximum without convergence.
In such instances, the corrupted codeword is retained as the decoding output for the calculation of BER and FER metrics. 

\subsubsection{Results on Correcting a Fixed Number of Errors}

\begin{table*}[tb!]
\centering
\caption{Performance of different VT decoders, including HD, SISO, VT-Former, and VT-Former-1, 
on correcting various numbers of errors. 
These models are tested on VT codes with code/message lengths $\mathrm{VT}(20,14)$, $\mathrm{VT}(68,60)$,
and $\mathrm{VT}(100,92)$.}
\begin{tabular}{rrcccccccccccc}
\toprule
 & & \multicolumn{4}{c}{$\mathrm{VT}(20,14)$} & \multicolumn{4}{c}{$\mathrm{VT}(68,60)$} & \multicolumn{4}{c}{$\mathrm{VT}(100,92)$} \\
\cmidrule(lr){3-6} \cmidrule(lr){7-10} \cmidrule(lr){11-14}
 & \# Error & 1 & 2 & 3 & 4 & 1 & 2 & 3 & 4 & 1 & 2 & 3 & 4 \\
\midrule
\multirow{2}{*}{\textbf{HD}} 
 & 1-BER (\%) & \textbf{100.0} & 68.9 & 65.2 & 60.5 & \textbf{100.0} & 68.1 & 65.9 & 61.0 & \textbf{100.0} & 68.0 & 65.3 & 61.2 \\
 & 1-FER (\%) & \textbf{100.0} & 12.5 & 8.9 & 2.4 & \textbf{100.0} & 4.5 & 2.7 & 0.3 & \textbf{100.0} & 2.0 & 1.7 & 0.1 \\
\midrule
\multirow{2}{*}{\textbf{SISO}} 
 & 1-BER (\%) & \textbf{100.0} & 83.1 & 74.3 & 73.6 & 99.9 & \textbf{87.3} & 81.0 & 77.9 & \textbf{100.0} & 83.2 & 80.2 & 79.1 \\
 & 1-FER (\%) & \textbf{100.0} & 13.8 & 4.9 & 0.7 & 99.4 & 4.2 & 1.6 & 0.2 & \textbf{100.0} & 1.4 & 0.9 & 0.1 \\
\midrule
\multirow{2}{*}{\textbf{VT-Former}} 
 & 1-BER (\%) & \textbf{100.0} & { 82.4} & 75.6 & 71.1 & \textbf{100.0} & 83.7 & 78.1 & 75.0 & \textbf{100.0} & 84.1 & {78.2} & 74.5 \\
 & 1-FER (\%) & \textbf{100.0} & {\textbf{30.1}} & \textbf{14.4} & \textbf{6.0} & \textbf{100.0} & \textbf{13.8} & \textbf{3.2} & \textbf{0.6} & \textbf{100.0} & \textbf{8.2} & \textbf{2.4} & \textbf{0.2} \\
\midrule
\multirow{2}{*}{\textbf{VT-Former-1}} 
             & 1-BER (\%) & \textbf{100.0} & \textbf{84.2} & \textbf{78.6} & \textbf{74.7} & \textbf{100.0} & 86.4 & {\textbf{81.6}} & {\textbf{78.5}} & \textbf{100.0} & \textbf{87.0} & \textbf{82.6} & {\textbf{79.9}} \\
         & 1-FER (\%) & \textbf{100.0} & 15.9 & 9.2 & 2.1 & \textbf{100.0} & {2.8} & 2.3 & {0.2} & \textbf{100.0} & {2.0} & 1.3 & {0.05} \\
\bottomrule
\end{tabular}
\label{tab:fixed-error}
\end{table*}

To quantitatively evaluate the performance of different decoding methods in multi-error correction scenarios, 
a fixed number of IDS errors (1, 2, 3, or 4) are artificially introduced into VT codewords of various lengths, 
where the insertion, deletion, and substitution errors are randomly distributed within the sequence.  
During the training phase of VT-Former for correcting $n$ errors, 
the model is exposed to samples containing a varying number of errors 
ranging from $0$ to $n$ in order to enhance its generalization capability across different number of errors. 
During the testing phase, its correction performance is evaluated under the condition 
that exactly $n$ errors are present in the codeword.  
The experimental results are summarized in \Cref{tab:fixed-error}.

As suggested in \Cref{tab:fixed-error}, when only a single IDS error is present, 
all compared methods achieve near-perfect correction performance, 
which is consistent with the theoretical ability of VT codes under the single-error assumption. 
However, as the number of errors increases, the performance of the HD and SISO decoders degrades rapidly. 
In contrast, the proposed VT-Former consistently maintains a clear advantage in terms of frame accuracy.  
Moreover, by incorporating a local window-based masking mechanism (with the window size set to $1$), 
VT-Former-1 also outperforms HD and SISO in terms of bit accuracy. 

These results indicate that, 
while VT codes are theoretically designed for single-error correction, 
partial information retrieval remains possible in multi-error scenarios. 
It is noteworthy that since VT codes are
``order optimal'' for single-error correction, 
the upper bound of the frame success rate ($1-\mathrm{FER}$) asymptotically approaches $50\%$ 
under two-error conditions as the code length increases. 
This observation further highlights the effectiveness of the proposed method.

Further analysis reveals a positive correlation between the decoding error rate and the number of errors, 
aligning with fundamental principles of channel coding theory. 
When varying the code lengths, it is observed that performance degradation caused by a fixed absolute number of errors 
becomes more pronounced as the code length increases. 
This observation suggests that longer codewords exhibit greater sensitivity to multi-error impairments, 
thereby underscoring the necessity of designing robust decoders specifically for multi-error environments. 

\subsubsection{Results on Independently Distributed Errors}
\begin{table}[tb!]
\centering
\caption{Performance of different VT decoders, including HD, SISO, VT-Former, and VT-Former-1, 
under various channels with different channel error rates. 
These models are tested on VT codes with code/message lengths $\mathrm{VT}(20,14)$, $\mathrm{VT}(68,60)$,
and $\mathrm{VT}(100,92)$.}
\setlength{\tabcolsep}{4 pt}
\begin{tabular}{rrcccccccc}
\toprule
 & & \multicolumn{2}{c}{$\mathrm{VT}(20,14)$} & \multicolumn{2}{c}{$\mathrm{VT}(68,60)$} & \multicolumn{2}{c}{$\mathrm{VT}(100,92)$} \\
\cmidrule(lr){3-4} \cmidrule(lr){5-6} \cmidrule(lr){7-8}
 & Error Rate & 1\% & 3\% & 1\% & 3\% & 1\% & 3\% \\
\midrule
\multirow{2}{*}{\textbf{HD}} 
 & 1-BER (\%) & 99.5 & 96.2 & 95.0 & 78.6 & 91.8 & 71.2 \\
 & 1-FER (\%) & 98.7 & 89.8 & 85.8 & 39.5 & 74.4 & 20.3 \\
 \midrule
 \multirow{2}{*}{\textbf{SISO}} 
 & 1-BER (\%) & \textbf{99.7} & 97.6 & 96.9 & 89.1 & 95.2 & 84.3 \\
 & 1-FER (\%) & 98.5 & 89.5 & 84.4 & 32.8 & 72.5 & 14.0 \\
\midrule
\multirow{2}{*}{\textbf{VT-Former}} 
 & 1-BER (\%) & \textbf{99.7} & 97.6 & 97.4 & 87.7 & 95.1 & 81.0 \\
 & 1-FER (\%) & \textbf{98.8} & \textbf{90.7} & \textbf{87.2} & \textbf{44.7} & \textbf{75.5} & \textbf{20.8} \\
\midrule
\multirow{2}{*}{\textbf{VT-Former-1}} 
 & 1-BER (\%) & \textbf{99.7} & \textbf{98.0} & \textbf{98.3} & \textbf{90.1} & \textbf{96.2} & \textbf{85.5} \\
 & 1-FER (\%) & 98.7 &89.9 & 86.7 & 41.2 & 74.3 & 17.9 \\
\bottomrule
\end{tabular}
\label{tab:probchannel}
\end{table}

In applications such as DNA storage, IDS errors typically occur in 
an approximately independent manner throughout the sequence, 
posing significant challenges to decoder robustness.  
To evaluate the performance of the proposed method under these channel conditions, 
IDS errors are independently introduced into the codewords.  
The channel error rates are set to $1\%$ and $3\%$, 
with the three error types occur with equal probability.  
The corresponding experimental results are summarized in \Cref{tab:probchannel}. 

For shorter codewords with low channel error rates, such as $\mathrm{VT}(20,14)$ at a $1\%$ channel error rate, 
the probability of multiple IDS errors 
occurring within a single codeword is relatively low. 
Under these conditions, 
all decoding methods are capable of reliably correcting single IDS errors, 
and thus the performance differences among the compared methods are not statistically significant. 

As the code length increases to $68$ or $100$, 
the probability of multiple IDS errors occurring within a single sequence rises.  
For $\mathrm{VT}(68,60)$ and $\mathrm{VT}(100,92)$, the proposed VT-Former demonstrates 
an advantage over both HD and SISO decoders in terms of frame-level performance. 
Furthermore, with the masking window size set to $1$, 
VT-Former-1 achieves superior bit-level decoding accuracy. 
These results indicate that the proposed method exhibits enhanced error-correction capability 
and superior robustness when addressing independently distributed multiple IDS errors. 

\subsection{Ablation study and hyperparameter optimization}

To evaluate the contributions of each component 
within the proposed model and to optimize key hyperparameter configurations, 
a detailed ablation study and hyperparameter sensitivity analysis are performed. 
Given that the utilized VT codes adhere to identical structural principles, 
these experiments are conducted on the $\mathrm{VT}(20,14)$ code by default,
unless otherwise specified. 

\subsubsection{Window size}
The window parameter $w$, defined in \Cref{eqn:windowmask}, determines the length of the local context accessible 
during the autoregressive decoding process, 
thereby governing the trade-off between exploiting local structures and capturing global dependencies. 
To assess the impact of this parameter, error correction performance is evaluated on corrupted $\mathrm{VT}(20,14)$ 
codewords across controlled error counts and varying values of $w$. 
The corresponding BER and FER are summarized in \Cref{tab:windowsize}. 

The results indicate that as the window size increases, 
a monotonic improvement in frame accuracy ($1-\mathrm{FER}$) is observed, 
whereas bit accuracy ($1-\mathrm{BER}$) exhibits a slight decline. 
This suggests that restricting symbol interaction to a limited local window, 
while beneficial for correcting local error patterns (as reflected in improved bit accuracy), 
may compromise the reconstruction of the global codeword structure (leading to reduced frame accuracy). 
Such observations demonstrate that effective error correction decoding necessitates both the 
fine-grained modeling of local context and the integration of long-range dependencies within the sequence. 
Consequently, reliance on a purely local window mechanism cannot fully substitute for the 
comprehensive modeling of global structural information.


\begin{table}[tb!]
\centering
\caption{Performance of VT-Former with different window sizes $w$ ranging from $1$ to code length $n$. 
The results are evaluated on $\mathrm{VT}(20,14)$ under different numbers of errors.}
\setlength{\tabcolsep}{3 pt}
\begin{tabular}{rrccccccc}
\toprule
\# Error &  & $w=1$ & $w=2$ & $w=5$ & $w=10$ & $w=15$ & $w=n$ \\
\midrule
\multirow{2}{*}{1} 
 & 1-BER (\%) & 100.0 & 100.0 & 100.0 & 100.0 & 100.0 & 100.0 \\
 & 1-FER (\%) & 100.0 & 100.0 & 100.0 & 100.0 & 100.0 & 100.0 \\
\midrule
\multirow{2}{*}{2} 
 & 1-BER (\%) & 84.2 & 82.8 & 82.3 & 82.5 & 82.4 & {82.4} \\
 & 1-FER (\%) & 15.9 & 19.9 & 23.1 & 28.7 & 29.6 & {30.1} \\
\midrule
\multirow{2}{*}{3} 
 & 1-BER (\%) & 78.6 & 77.3 & 76.6 & 75.9 & 75.8 & 75.6 \\
 & 1-FER (\%) & 9.2 & 10.0 & 11.2 & 13.7 & 14.1 & 14.4 \\
\midrule
\multirow{2}{*}{4} 
 & 1-BER (\%) & 74.7 & 73.2 & 72.8 & 71.6 & 71.3 & 71.1 \\
 & 1-FER (\%) & 2.1 & 2.8 & 3.7 & 5.2 & 5.8 & 6.0 \\
\bottomrule
\end{tabular}
\label{tab:windowsize}
\end{table}

\subsubsection{Statistic Embedding}
\begin{table}[tb!]
\caption{
Performance of different statistic embedding strategies on $\mathrm{VT}(20,14)$ under varying numbers of IDS errors. 
The notation A/B indicates the presence (w) or absence (wo) of statistic (embedding) 
for the corrupted and predicted codewords, respectively.
}
\centering
\begin{tabular}{rrcccc}
\toprule
\# Error &  & w/w & w/wo & wo/w& wo/wo \\
\midrule
\multirow{2}{*}{1} 
 & 1-BER (\%) & 100.0 & 87.3 &86.1& 85.8 \\
 & 1-FER (\%) & 100.0 & 7.3  &11.6& 12.1 \\
\midrule
\multirow{2}{*}{2} 
 & 1-BER (\%) & {82.4} & 77.9 &78.6& 78.3 \\
 & 1-FER (\%) & {30.1 }& 2.3  &2.0& 1.9  \\
\midrule
\multirow{2}{*}{3} 
 & 1-BER (\%) & 75.6 & 73.3 &75.0& 74.7 \\
 & 1-FER (\%) & 14.4 & 1.7  &1.2& 1.3  \\
\midrule
\multirow{2}{*}{4} 
 & 1-BER (\%) & 71.1 & 70.9 &71.8& 72.0 \\
 & 1-FER (\%) & 6.0  & 0.5  &0.2& 0.3  \\
\bottomrule
\end{tabular}
\label{tab:statistic}
\end{table}

The statistic embedding of the codeword, introduced as part of the codeword embedding in \Cref{sec:symbol-statistic}
is evaluated in ablation studies, as shown in \Cref{tab:statistic}. 
In this table, the column headers indicate whether the statistic embedding is applied to the embedding of 
the corrupted codeword and the predicted codeword. 
For instance, the notation ``\textbf{wo/w}'' indicates that the corrupted codeword is processed \textbf{without} statistic embedding, 
whereas the model is trained to reconstruct the codeword \textbf{with} statistic, 
thereby necessitating the explicit prediction of these statistic features during the decoding process. 

The results in \ref{tab:statistic} suggest that introducing the statistic embedding into 
the input codeword embedding significantly improves frame accuracies ($1$-FER). 
Meanwhile, in the case of ``wo/wo'', where the statistic embedding is absent, 
the model's performance decreases considerably. 

Specifically, the application of the statistic embedding strategy yields the most prominent improvement 
in the single-error scenario, elevating the model's correction capability to $100\%$. 
Given that VT codes are mathematically designed to correct single IDS errors and that HD decoding exploits 
statistical constraints to achieve this, 
it is postulated that the proposed embedding effectively captures and aligns with the combinatorial properties of the VT code. 
However, as the error count increases, the performance disparity between the different embedding strategies diminishes. 
This indicates that the utilized statistic information becomes less effective in guiding the correcting process 
for multiple errors, rendering the gains from statistic embedding less pronounced.

\subsubsection{Decoder-only VT-Former}

Conventional sequence-to-sequence models typically comprise both encoder and decoder stacks. 
To enhance computational efficiency, the corrupted codeword embeddings could be utilized directly as 
the encoder memory for cross-attention, thereby bypassing the computational overhead of a dedicated encoder.

Experiments are conducted to assess whether this streamlined configuration compromises 
decoding performance while improving efficiency. 
The resulting decoder-only architecture, designated as VT-Former-D, 
employs a 3-layer decoder to predict the ground-truth codeword directly from the embeddings. 
The corresponding results are presented in \Cref{tab:encoder}.

Experimental results suggested that VT-Former-D exhibits a similar performance  
in both FER and BER, compared to the full VT-Former model. 
This indicates that the decoder component of the model is capable of effectively leveraging 
the codeword embeddings to perform accurate decoding. 

\begin{table*}[tb!]
\centering
\caption{Performance of different VT decoders, comparing the decoder-only architecture VT-Former-D with 
the VT-Former and VT-Former-1 model.}
\begin{tabular}{rrcccccccccccc}
\toprule
 & & \multicolumn{4}{c}{$\mathrm{VT}(20,14)$} & \multicolumn{4}{c}{$\mathrm{VT}(68,60)$} & \multicolumn{4}{c}{$\mathrm{VT}(100,92)$} \\
\cmidrule(lr){3-6} \cmidrule(lr){7-10} \cmidrule(lr){11-14}
 & \# Error & 1 & 2 & 3 & 4 & 1 & 2 & 3 & 4 & 1 & 2 & 3 & 4 \\
\midrule
\multirow{2}{*}{\textbf{VT-Former}} 
 & 1-BER (\%) & \textbf{100.0} & {82.4} & 75.6 & 71.1 & \textbf{100.0} & 83.7 & 78.1 & 75.0 & \textbf{100.0} & 84.1 & {78.2} & 74.5 \\
 & 1-FER (\%) & \textbf{100.0} & {\textbf{30.1}} & \textbf{14.4} & \textbf{6.0} & \textbf{100.0} & \textbf{13.8} & \textbf{3.2} & \textbf{0.6} & \textbf{100.0} & \textbf{8.2} & \textbf{2.4} & \textbf{0.2} \\
\midrule
\multirow{2}{*}{\textbf{VT-Former-1}} 
 & 1-BER (\%) & \textbf{100.0} & \textbf{84.2} & \textbf{78.6} & \textbf{74.7} & \textbf{100.0} & \textbf{86.4} & {\textbf{81.6}} & \textbf{78.5} & \textbf{100.0} & \textbf{87.0} & \textbf{82.6} & \textbf{79.9} \\
 & 1-FER (\%) & \textbf{100.0} & 15.9 & 9.2 & 2.1 & \textbf{100.0} & {2.8} & 2.3 & {0.2} & \textbf{100.0} & {2.0} & {1.3} & {0.05} \\
\midrule
 \multirow{2}{*}{\textbf{VT-Former-D}} 
 & 1-BER (\%) & \textbf{100.0} & 82.3 & 75.6 & 72.8 & \textbf{100.0} & 82.3 & 77.7 & 74.7 & \textbf{100.0} & 83.0 & {78.0} & 74.8 \\
 & 1-FER (\%) & \textbf{100.0} & 28.2 & 13.2 & 6.9 & \textbf{100.0} & 10.8 & 3.0 & 0.5 & \textbf{100.0} & 6.5 & 2.4 & {0.1} \\
\bottomrule
\end{tabular}
\label{tab:encoder}
\end{table*}

\subsubsection{Performance analysis under different error types}

\begin{figure*}[tb!]
    \centering
    \subfloat[Bit accuracy.]{
        \includegraphics[width=0.32\linewidth]{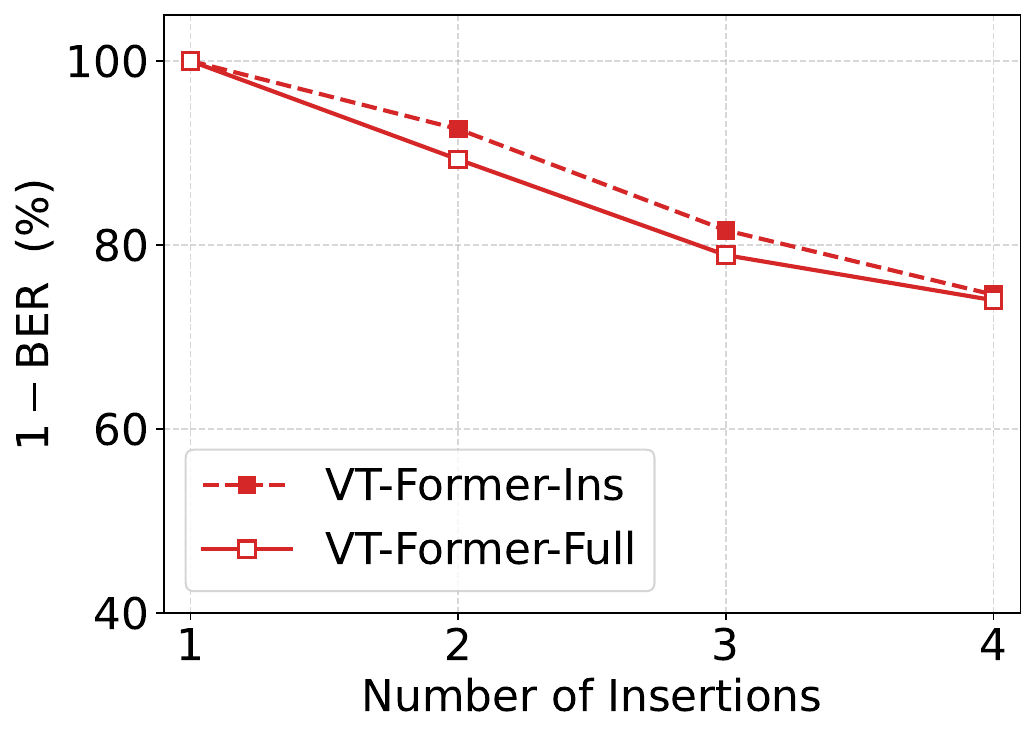}
        \includegraphics[width=0.32\linewidth]{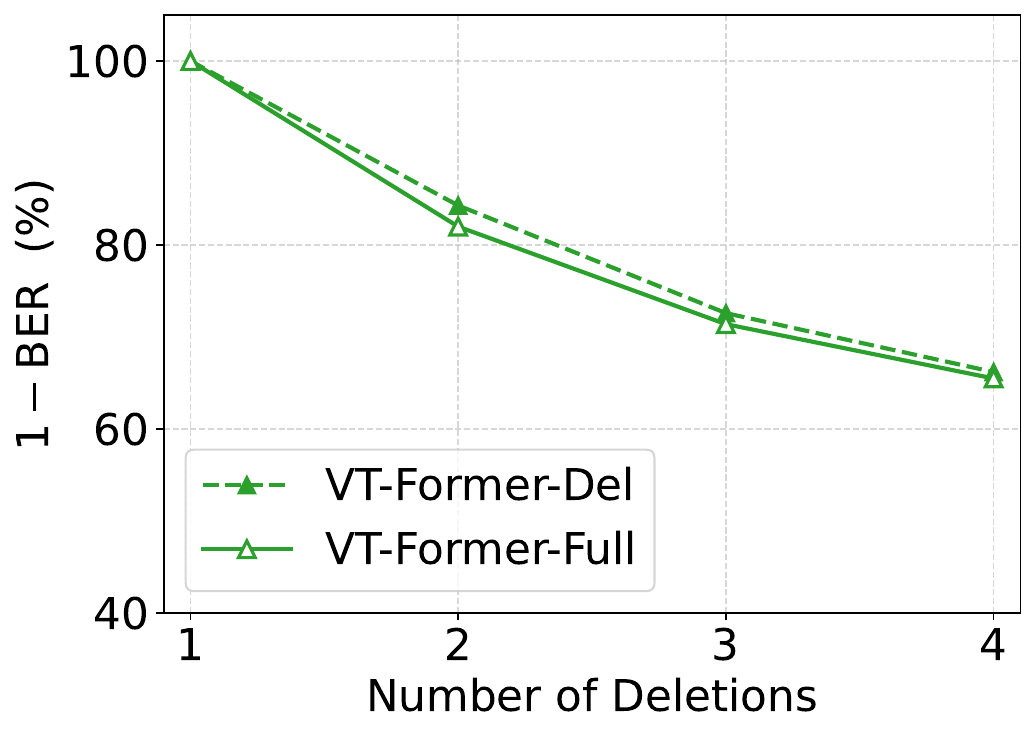}
        \includegraphics[width=0.32\linewidth]{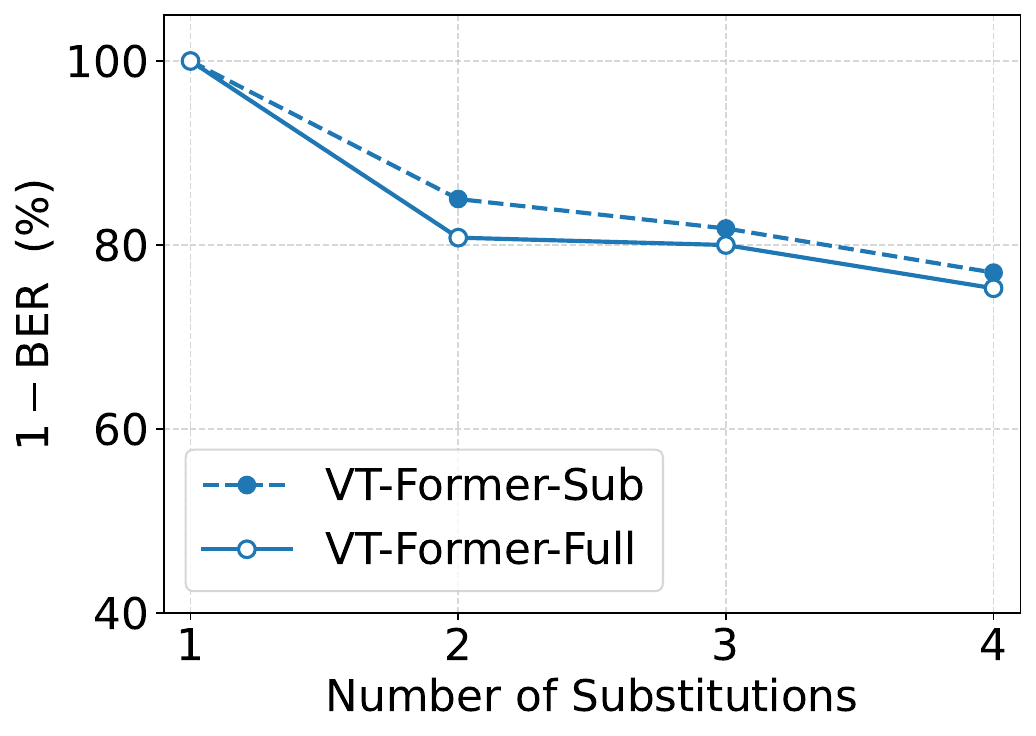}
        }\\
    \subfloat[Frame accuracy.]{
        \includegraphics[width=0.32\linewidth]{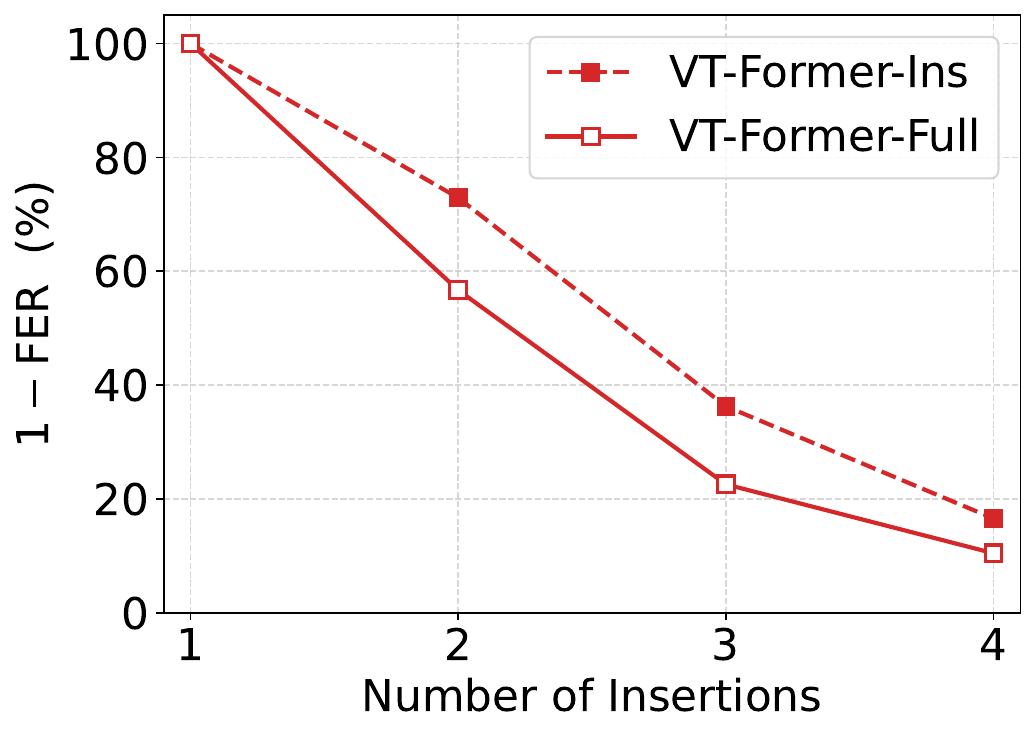}
        \includegraphics[width=0.32\linewidth]{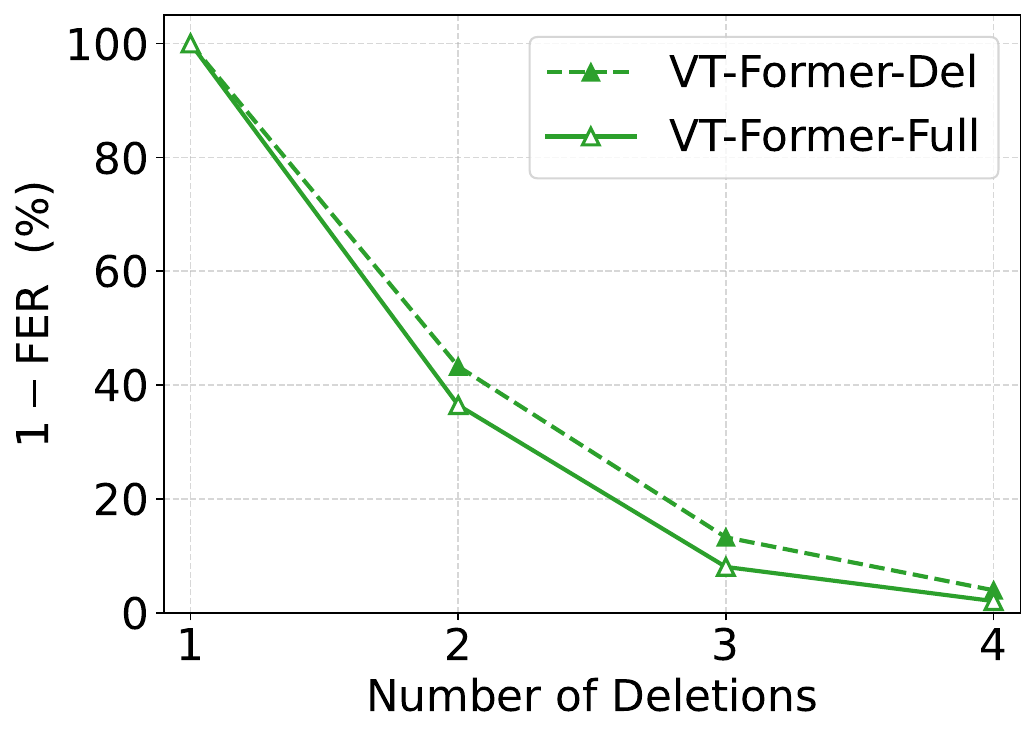}
        \includegraphics[width=0.32\linewidth]{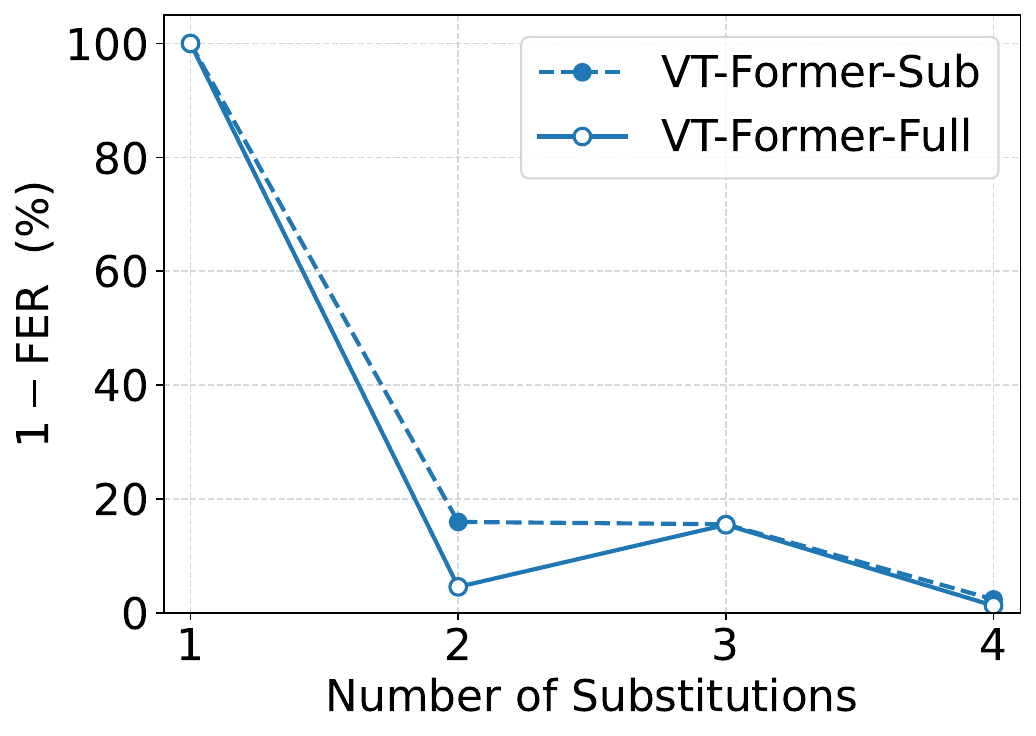}
        }
    \caption{Bit accuracy and frame accuracy of the VT-Former architecture 
    on correcting specific error types, including insertion, deletion, and substitution errors. 
    The suffixes denote models trained on the corresponding single error type or 
    the full mixed-error dataset, respectively.\label{fig:ber-fer}}
\end{figure*}

Insertion, deletion, and substitution errors exhibit distinct characteristics in 
their impact on sequence structure and content integrity: 
insertion errors primarily introduce redundant symbols, 
deletion errors cause positional shifts in the sequence, 
while substitution errors directly alter symbol content. 

To evaluate the error correction capabilities of VT-Former regarding specific error types, 
experiments are conducted wherein codewords are subjected to a single type of error. 
For each error type, evaluation is performed across error counts ranging from $1$ to $4$. 
Additionally, the study contrasts two training paradigms: 
models trained exclusively on a specific error type versus those trained on the full spectrum of errors. 
In \Cref{fig:ber-fer}, the suffix of error type denotes a model trained solely on the corresponding single error type, 
while the suffix ``Full'' indicates a model trained on the complete dataset encompassing all error types.

As illustrated in \Cref{fig:ber-fer}, 
a consistent decline in accuracy is observed across all metrics as the number of errors increases, 
aligning with theoretical expectations. 
Regarding bit accuracy, the VT-Former exhibits comparable performance for insertion and substitution errors, 
whereas deletion errors result in comparatively lower accuracy. 
In terms of frame accuracy, the model demonstrates superior resilience to insertion errors 
compared to the other two types. 
It is postulated that, although insertion errors disrupt the sequence structure, 
they preserve the original symbols in the sequence, 
thereby facilitating the recovery of the correct codeword. 
A comparison between the dashed and solid curves reveals 
that models trained on specific error types generally outperform those trained on the full mixed-error dataset, 
a result consistent with anticipation. 

\subsection{Complexity Analysis}
\begin{figure*}[tb!]
    \centering
    \subfloat[Decoding throughput in terms of codewords per second.]{
        \includegraphics[width=0.4\linewidth]{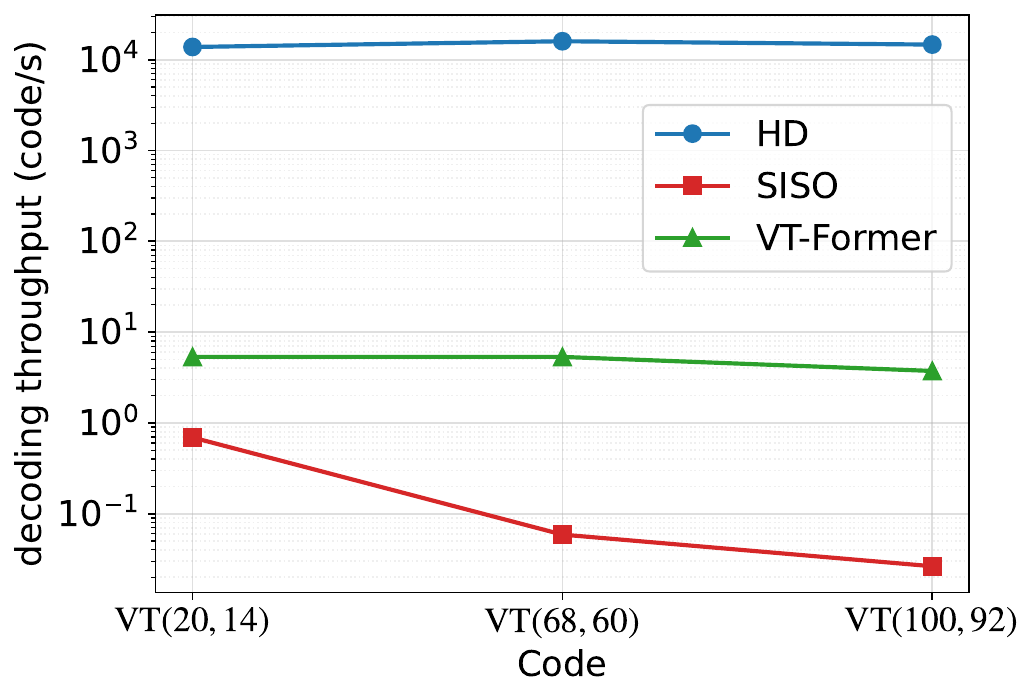}
        }
        \quad
    \subfloat[Decoding throughput in terms of bits per second.]{
        \includegraphics[width=0.4\linewidth]{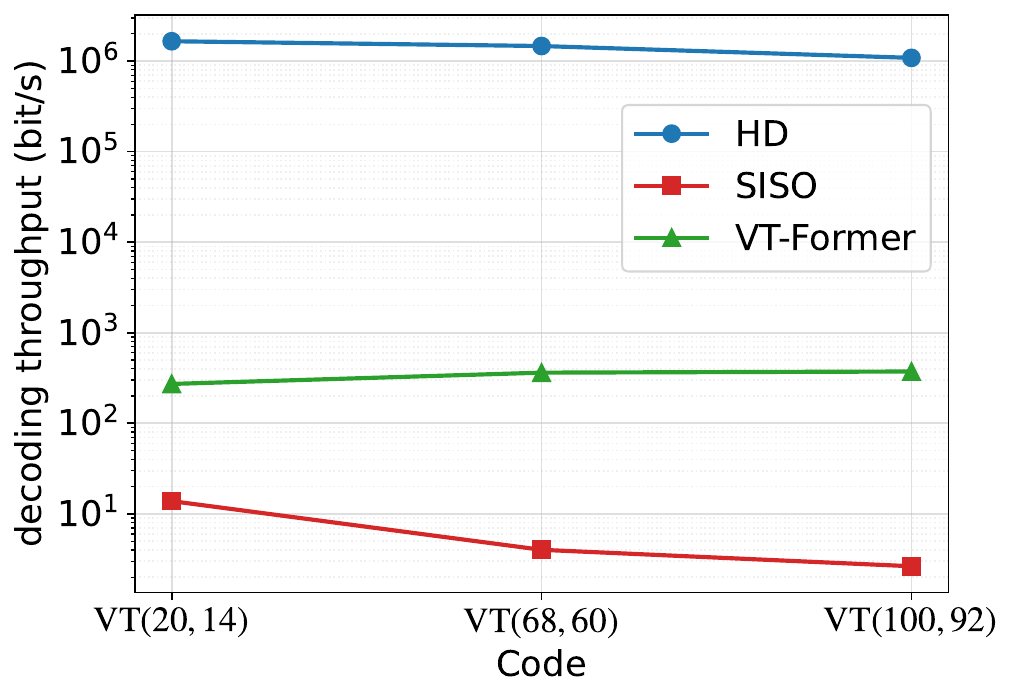}
        }
    \caption{Decoding throughput in terms of codewords per second and bits per second 
    across different decoders and code lengths.\label{fig:decoding_throughput}}
\end{figure*}
The computational complexity of the VT-Former is primarily attributed to three core components: 
the cross-attention mechanism, the self-attention mechanism, and the feed-forward network. 
Let $n$ denote the code length and $d$ denote the embedding dimension.  
The complexity of the cross-attention mechanism is $O(n^2 d)$, 
while the self-attention mechanism, constrained by a local window size $w$, scales as $O(n w d)$, 
and the feedforward network contributes a complexity of $O(n d^2)$. 
Given that the window size $w$ is upper-bounded by the code length ($w \leq n$), 
the overall complexity of the decoder is dominated by the attention mechanisms, resulting in a total complexity of $O(n^2 d)$.

Although the standard Transformer exhibits a quadratic complexity of $O(n^2 d)$, 
VT-Former reduces decoding time by leveraging GPU parallelism. 
On a machine equipped with two Intel Xeon Silver 4116 CPU and an NVIDIA RTX 3090 GPU, 
we evaluated the time consumption of the HD decoder, SISO decoder, and VT-Former on 
correcting single IDS errors in codewords from $\mathrm{VT}(20,14)$, $\mathrm{VT}(68,60)$, and $\mathrm{VT}(100,92)$. 
The HD and SISO decoders are implemented on the CPU, while VT-Former is implemented on the GPU. 
The results on decoding throughput in terms of codewords per second 
and bits per second are illustrated in \Cref{fig:decoding_throughput}. 


The experimental results indicate that the HD decoder, functioning as a combinatorial decoder, 
achieves the shortest decoding latency and the highest throughput. 
In contrast, the SISO decoder exhibits the longest decoding time and the lowest throughput. 
This significant computational overhead is primarily attributed to the iterative implementation 
of the BCJR and MAP algorithms on the CPU. 
It is noteworthy that the HD decoder is theoretically limited to correcting a single error; 
consequently, its performance degrades precipitously in multi-error scenarios. 
Similarly, the decoding efficiency of the SISO decoder further diminishes in multi-error contexts 
compared to single-error cases, 
as the inference process faces convergence challenges, 
thereby necessitating a greater number of iterations.

Leveraging advanced GPU acceleration, 
the VT-Former achieves significantly lower decoding latency compared to the SISO decoder, 
while simultaneously maintaining superior multi-error correction capabilities. 
Specifically, the proposed method yields a throughput several orders of magnitude higher than the SISO baseline. 
The observed variations in decoding efficiency across different code lengths are primarily attributed 
to the dynamics of GPU parallelism and memory bandwidth. 
As suggested in \Cref{fig:decoding_throughput}, the VT-Former strikes a favorable balance between accuracy 
and computational efficiency.




\section{Conclusion}
This paper proposes VT-Former, the first Transformer-based decoder designed for VT codes.
By integrating symbol embedding with statistic feature representation and 
introducing a combined masking strategy, VT-Former achieves efficient joint correction of insertion, 
deletion, and substitution errors. 
Experimental results demonstrate that across various code lengths and error scenarios, 
VT-Former significantly reduces decoding error rates compared to conventional HD and SISO decoders. 
Furthermore, ablation studies confirm the critical role of statistic embedding 
and the local window-based masking mechanism in enhancing decoding performance. 
This work not only explores the theoretical boundaries of information retrieval for VT codes in multi-error scenarios 
but also provides a viable decoding solution with practical deployment potential for 
high-fault-tolerance applications, such as DNA-based information storage. 

\bibliographystyle{IEEEtran}
\bibliography{ref}
\newpage

\vfill

\end{document}